\documentclass{llncs}
\title{Exploiting Resolution-based Representations\\for MaxSAT Solving}
\usepackage{amssymb}
\usepackage[utf8]{inputenc}
\usepackage[T1]{fontenc}
\usepackage{comment}
\usepackage[linesnumbered,ruled,vlined]{algorithm2e}
\usepackage{graphicx}
\usepackage{tikz}
\usepackage{subfigure}

\definecolor{citeblue}{rgb}{0.1,0,.4}
\usepackage[pdftex%
,colorlinks=true%
,bookmarks=false%
,linkcolor=citeblue%
,citecolor=citeblue%
,urlcolor=blue%
,plainpages=false]{hyperref}
\hypersetup{pdfauthor={Neves et al.}%
,pdftitle={Exploiting Resolution-based Representations for MaxSAT Solving}
}

\begin{document}

\author{Miguel Neves\inst{1} \and Ruben Martins\inst{2} \and Mikol\'a\v{s} Janota\inst{1} \and \\ 
In\^es Lynce\inst{1} \and Vasco Manquinho\inst{1}}
\authorrunning{Neves, Martins, Janota, Lynce and Manquinho} 
\tocauthor{Miguel Neves, Ruben Martins, Mikolas Janota, Ines Lynce, Vasco Manquinho}
\institute{INESC-ID / Instituto Superior T\'ecnico, Universidade de Lisboa, Portugal\\
  \email{\{neves,mikolas,ines,vmm\}@sat.inesc-id.pt}
  \and
  University of Oxford, Department of Computer Science, United Kingdom\\
  \email{ruben.martins@cs.ox.ac.uk}}

\maketitle

\begin{abstract}
  Most recent MaxSAT algorithms rely on a succession of
  calls to a SAT solver in order to find an optimal solution. In particular,
  several algorithms take advantage of the ability of SAT solvers to
  identify unsatisfiable subformulas. Usually, these MaxSAT algorithms 
  perform better when small unsatisfiable subformulas are found early. 
  However, this is not the case in many problem instances, since the whole 
  formula is given to the SAT solver in each call.
  
  In this paper, we propose to partition the MaxSAT formula using a 
  resolution-based graph representation. Partitions are then iteratively 
  joi\-ned by using a proximity measure extracted from the graph representation 
  of the formula. The algorithm ends when only one partition remains and the 
  optimal solution is found.
  Experimental results show that this new approach further enhances a state
  of the art MaxSAT solver to optimally solve a larger set of industrial
  problem instances.
\end{abstract}


\section{Introduction}
\label{sec:intro}

Many real-world problems in different areas such as fault localization in 
C programs, design debugging, upgradability of software systems, among other,
can be solved using Maximum Satisfiability 
(MaxSAT)~\cite{asin2012curriculum,PackUp12,majumdar-pldi11,le2010dependency,veneris-fmcad07}. 
The increase usage of MaxSAT for problem solving results
from the improvements of MaxSAT technology in recent years.
In the last decade, several new techniques and algorithms have been
proposed that improved on previous MaxSAT solvers by several orders of
magnitude. Moreover, the developments in the underlying SAT technology,
namely identification of unsatisfiable subformulas and incrementality
have also been a factor in the  improvements of MaxSAT solving.

MaxSAT solvers for industrial instances are usually based on iterative calls 
to a SAT solver. Moreover, most of these MaxSAT algorithms take advantage 
of the ability of SAT solvers to identify unsatisfiable subformulas. 
However, in most cases, algorithms deal with the whole formula at each
call of the SAT solver. As a result, unnecessarily large unsatisfiable
subformulas can be returned at each SAT call, resulting in a slow down
of the MaxSAT algorithm. In this work, we try to avoid this behavior
by partitioning the formula and taking advantage of structural information
obtained from a formula's graph representation.

In this paper, we improve on the current state of the art MaxSAT
solving by proposing a new unsatisfiability-based algorithm for
MaxSAT. The new algorithm integrates several new features,
namely: (1)~usage of resolution-based graphs to represent the
MaxSAT formula, (2)~partition of soft clauses in the MaxSAT
formula using the referred representation, (3)~usage of structural
information obtained from the graph representation to drive
the merge of partitions and, (4)~integration of these features
into a new fully incremental algorithm that improves on the best
non-portfolio solver from the last MaxSAT Solver Evaluation
on several industrial benchmark sets.


The paper is organized as follows. Section~\ref{sec:prelim} formally
defines MaxSAT and briefly reviews the MaxSAT algorithms more closely
related to the proposed approach. In section~\ref{sec:res}, graph
representations of CNF formulas are described. Moreover, the adaptation
of resolution-based graphs is proposed. The new MaxSAT algorithm is
proposed in section~\ref{sec:msu3-new}. Besides a detailed description,
we show how to extract structural information from the graph representations
and integrate it in the new algorithm. Section~\ref{sec:results} presents
the experimental results of the new MaxSAT solver on a large set of
industrial benchmark sets used at MaxSAT evaluations. Finally, the
paper concludes in section~\ref{sec:conc}.

\section{Preliminaries}
\label{sec:prelim}

A propositional formula in Conjunctive Normal Form (CNF), using 
$n$ Boolean variables $x_1 , x_2 , \ldots , x_n$, is defined
as a conjunction of clauses, where a clause is a disjunction
of literals. A literal is either a variable $x_i$ or its complement 
$\bar x_i$.
The Propositional Satisfiability (SAT) problem consists of deciding 
whether there exists a truth assignment to the variables such that 
the formula is satisfied.

The Maximum Satisfiability (MaxSAT) can be seen as an optimization version
of the SAT problem. In MaxSAT, the objective 
is to find an assignment to the variables of a CNF formula that
minimizes the number of unsatisfied clauses. 
Notice that minimizing the number of unsatisfied clauses is equivalent 
to maximizing the number of satisfied clauses.

In a partial MaxSAT formula $\varphi = \varphi_h \cup \varphi_s$, some
clauses are considered as hard ($\varphi_h$), while others are declared
as soft ($\varphi_s$). The goal in partial MaxSAT is to find an assignment
to the formula variables such that all hard clauses in $\varphi_h$ are satisfied, 
while minimizing the number of unsatisfied soft clauses in $\varphi_s$.
There are also weighted variants of MaxSAT where soft clauses are
associated with weights greater than or equal to 1. In this case, the
objective is to satisfy all hard clauses and minimize the total weight
of unsatisfied soft clauses. In this paper, we focus solely on partial
MaxSAT, but the proposed approach can be generalized to its weighted
variants. Furthermore, in all algorithms we assume that the set of hard
clauses $\varphi_h$ is satisfiable. Otherwise, the MaxSAT formula
does not have a solution. This can easily be checked through a
SAT call on $\varphi_h$.


The most recent state of the art MaxSAT solvers are based on
iterative calls to a SAT solver. One of the most classic
approaches is the linear Sat-Unsat algorithm that performs a 
linear search on the number of unsatisfied clauses. 
In this case, a new relaxation variable
is initially added to each soft clause and the resulting
formula is given to a SAT solver. Whenever a solution is
found, a new cardinality constraint on the number of relaxation 
variables is added, such that solutions where a higher or equal 
number of relaxation variables assigned the value~1 are excluded. 
The cardinality constraint is encoded into a set of propositional
clauses, which are added to the working
formula~\cite{bailleux-cp03,qmaxsat-jsat12,martins-cp14}.
The algorithm stops when the SAT call is unsatisfiable. As a result,
the last solution found is an optimal solution of the MaxSAT formula.

\DontPrintSemicolon
\SetKwFunction{soft}{soft}
\SetKwFunction{SAT}{SAT}
\SetKwFunction{decomposeSoft}{partitionSoft}
\SetKwFunction{first}{first}
\SetKwFunction{weight}{weight}
\SetKwFunction{encodeCNF}{CNF}
\SetKwFunction{min}{min}
\SetKwFunction{refineBound}{UpdateBound}
\SetKwData{result}{satisfiable assignment to}
\SetKwData{unsat}{UNSAT}
\SetKwData{sat}{SAT}
\SetKwData{minc}{min$_\textnormal{c}$}
\SetKwData{true}{true}
\SetKwData{st}{st}
\SetVlineSkip{1pt}
\begin{algorithm}[!t]
  \small
  \KwIn{$\varphi = \varphi_h \cup \varphi_s$}
  \KwOut{satisfying assignment to $\varphi$}
  $(\varphi_W, V_R, \lambda) \gets (\varphi_h, \emptyset, 0)$\;
  \ForEach{$\omega_i \in \varphi_s$}{
    $V_R \gets V_R \cup \{ r_i \}$\tcp*[r]{\footnotesize $r_i$ is a new relaxation variable}
    $\omega_R \gets \omega_i \cup \{ r_i\}$\;
    $\varphi_W \gets \varphi_W \cup \{ \omega_R \}$\;
  }
  \While{\true}{
    $(\st, \nu, \varphi_C) \gets \SAT(\varphi_W \cup \{ \encodeCNF( \sum_{r_i\in V_R} r_i \leq \lambda ) \})$ \label{linearus:satcall}\;
    \If{$\st = \sat$}{
      \Return{$\nu$}\tcp*[r]{\footnotesize satisfying assignment to $\varphi$}\label{linearus:optFound}
    } 
    $\lambda \gets \lambda + 1$\label{linearus:inclb}\;
  }
  \caption{Linear Search Unsat-Sat Algorithm}\label{alg:linear-us}
\end{algorithm}

A converse approach is the linear search Unsat-Sat presented in 
Algorithm~\ref{alg:linear-us}.
Here, a lower bound $\lambda$ is maintained between iterations
of the algorithm. Initially, $\lambda$ is assigned value 0.
In each iteration, while the working formula given to the SAT solver 
(line~\ref{linearus:satcall}) is unsatisfiable, $\lambda$ is 
incremented (line~\ref{linearus:inclb}).
Otherwise, an optimal solution to the MaxSAT formula has been 
found (line~\ref{linearus:optFound}).

Observe that a SAT solver call on a CNF formula $\varphi_W$ returns a 
triple ({\em st}, $\nu$, $\varphi_C$), where {\em st} denotes the 
status of the solver: satisfiable (\textsf{SAT}) or 
unsatisfiable (\textsf{UNSAT}). If $\varphi_W$ is satisfiable, then 
$\nu$ stores the model found for $\varphi_W$. Otherwise, $\varphi_C$
contains an unsatisfiable subformula that explains a reason for the 
unsatisfiability of $\varphi_W$.

\DontPrintSemicolon
\SetKwFunction{soft}{soft}
\SetKwFunction{SAT}{SAT}
\SetKwFunction{decomposeSoft}{partitionSoft}
\SetKwFunction{first}{first}
\SetKwFunction{weight}{weight}
\SetKwFunction{encodeCNF}{CNF}
\SetKwFunction{min}{min}
\SetKwData{result}{satisfiable assignment to}
\SetKwData{unsat}{UNSAT}
\SetKwData{sat}{SAT}
\SetKwData{minc}{min$_\textnormal{c}$}
\SetKwData{true}{true}
\SetKwData{st}{st}
\SetVlineSkip{1pt}
\begin{algorithm}[!t]
  \small
  \KwIn{$\varphi = \varphi_h \cup \varphi_s$}
  \KwOut{satisfying assignment to $\varphi$}
  $(\varphi_W, V_R, \lambda) \gets (\varphi, \emptyset, 0)$\;
  \While{\true}{
    $(\st, \nu, \varphi_C) \gets \SAT(\varphi_W \cup \{ \encodeCNF( \sum_{r_i \in V_R} r_i \leq \lambda ) \})$ \label{msu3:satcall}\;
    \If{$\st = \sat$}{
      \Return{$\nu$}\tcp*[r]{\footnotesize satisfying assignment to $\varphi$}
    }
    \ForEach{$\omega_i \in (\varphi_C ~\cap~ \varphi_s)$}{
       $V_R \gets V_R \cup \{ r_i \}$\tcp*[r]{\footnotesize $r_i$ is a new variable}
       $\omega_R \gets \omega_i \cup \{ r_i \}$\tcp*[r]{\footnotesize $\omega_i$ was not previously relaxed}
       $\varphi_W \gets (\varphi_W \setminus \{ \omega_i \}) \cup \{ \omega_R \}$\;
    }
    $\lambda \gets \lambda + 1$\;
  }
  \caption{MSU3 Algorithm}\label{alg:msu3}
\end{algorithm}

Several of the most effective algorithms for MaxSAT take advantage
of the current SAT solvers being able to produce certificates
of unsatisfiability. Since the SAT solver is able to
identify unsatisfiable subformulas, several MaxSAT algorithms
use it to delay the relaxation of soft clauses.
An example is the MSU3 algorithm~\cite{wmsu3-corr07} presented in 
Algorithm~\ref{alg:msu3}. Observe that this algorithm
also performs an Unsat-Sat linear search, but soft clauses
are only relaxed when they appear in an unsatisfiable
subformula.

Although more sophisticated MaxSAT algorithms exist~\cite{morgado-constraints13},
an implementation of MSU3 algorithm on the Open-WBO framework
was the best performing non-portfolio algorithm at the
MaxSAT Solver Evaluation in 2014\footnote{Results available at 
\url{http://www.maxsat.udl.cat/}}. One of the crucial features
for its success relies on the fact that only one SAT solver
instance needs to be created~\cite{martins-cp14}. Therefore, 
a proper implementation of MSU3 should take advantage of incrementality
in SAT solver technology. In this paper, the MSU3 algorithm
is further improved with structural information of the problem
instance to solve.

\section{Graph Representations}
\label{sec:res}

In order to extract structural properties of CNF formulas,
different graph-based models have been previously proposed. 
For instance, graph representations have been used to
characterize industrial SAT instances~\cite{ansotegui-sat12}
and to improve on the performance of MaxSAT algorithms~\cite{martins-sat13}.
In this section, we briefly review the Clause-Variable Incidence 
Graph (CVIG) and adapt the use of Resolution-based Graphs (RES)~\cite{yates-AI70}
to model relations in CNF formulas. Although other models 
exist~\cite{ansotegui-sat12,vangelder-cp11,martins-sat13},
in the context of our algorithm for MaxSAT solving, these were
found to be the best suited.

In the CVIG model, a weighted undirected graph $G$ is built such that a 
vertex is added for each variable $x_j$ and for each clause $\omega_i$ 
occurring in the CNF formula~$\varphi$. 
Moreover, for each variable $x_j$ occurring in clause $\omega_i$ (either
as literal $x_j$ or $\bar x_j$), an edge $(\omega_i, x_j)$ is added 
to graph $G$. The edge weight $w(\omega_i, x_j)$ is defined as:
\begin{equation}
  w(\omega_i, x_j) = \frac{I(x_j)}{|\omega_i|}
\end{equation}
where $|\omega_i|$ denotes the number of literals in clause $\omega_i$
and $I(x_j)$ is defined as the incidence function of $x_j$ in soft
clauses as:
\begin{equation}
  I(x_j) = 1 + \sum_{x_j \in \omega \mbox{ } \wedge \mbox{ } \omega \in \varphi_s} \frac{1}{|\omega|}
\end{equation}


As described in section~\ref{sec:prelim}, several MaxSAT solvers rely on the 
identification of unsatisfiable subformulas. In order to capture sets
of clauses more closely related that would result in an unsatisfiable subformula,
we propose to adapt Resolution Graphs (RES) to MaxSAT.

In the RES model, we have one vertex in graph $G$ for each clause 
$\omega_i \in \varphi$. Let $\omega_i$ and $\omega_j$ denote two clauses 
such that $x_k \in \omega_i$ and $\bar x_k \in \omega_j$.
Moreover, let $\omega^{res}_{ij}$ be the resulting clause of applying the 
resolution operation on these clauses. In this case, if $\omega^{res}_{ij}$ 
is not a tautology, then an edge $(\omega_i, \omega_j)$ is added to $G$ 
whose weight is defined as:
\begin{equation}
  w(\omega_i, \omega_j) = \frac{1}{|\omega^{res}_{ij}|}
\end{equation}

Notice that in the RES model, clauses are related if the application
of the resolution operation results in a non-trivial resolvent.
Moreover, observe that the weight of edges between pairs of clauses
is greater when the size of the resolvent is smaller. The goal is
to make tighter the relations between clauses that produce smaller
clauses when resolution is applied.

\begin{figure}[t]
  \centering
  \subfigure[CVIG Graph]{
    \begin{tikzpicture}[thick,xscale=.4]
  \tikzstyle{every node}=[circle,draw]
  \foreach \a in {1,2,3} \node (x\a) at (\a*6-6,2)  {$x_{\a}$} ;
  \foreach \a in {1,...,5} \node (w\a) at (\a*3-3,0)  {$\omega_{\a}$} ;
  \foreach \i in {1,3,4} \draw(x1)--(w\i);
  \foreach \i in {1,2} \draw(x2)--(w\i);
  \foreach \i in {2,3,5} \draw(x3)--(w\i);
\end{tikzpicture}
    \label{fig:cvig}
  }
  \hspace{2em}
  \subfigure[RES Graph]{
    \begin{tikzpicture}[thick,xscale=1.3]
  \tikzstyle{every node}=[circle,draw]
  \node(w4) at (0,2) {$\omega_4$};
  \node(w1) at (1,2) {$\omega_1$};
  \node(w3) at (2,2) {$\omega_3$};
  \node(w5) at (0,0) {$\omega_5$};
  \node(w2) at (1,0) {$\omega_2$};
  \draw(w4)--(w1)--(w3)--(w2)--(w5);
  \draw(w1)--(w2);
\end{tikzpicture}
    \label{fig:res}
  }
  \caption{Example of Graph Models}
\end{figure}
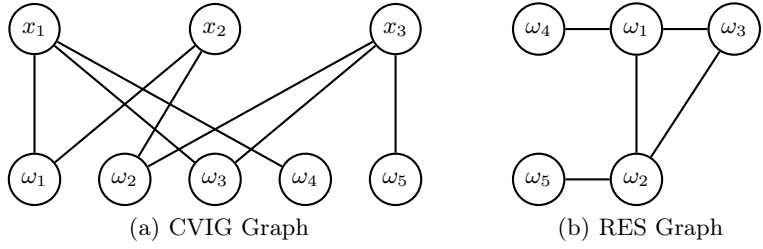

Consider the following MaxSAT formula where $\omega_1: (x_1 \vee x_2)$,
$\omega_2: (\bar x_2 \vee x_3)$ and $\omega_3: (\bar x_1 \vee \bar x_3)$
are hard clauses and $\omega_4: (\bar x_1)$, $\omega_5: (\bar x_3)$ are
soft clauses. Figures~\ref{fig:cvig} and \ref{fig:res} illustrate the
structure of the graph representation of this formula when using the
CVIG and RES models. The weights of edges are not represented for simplicity.
Observe that if the clause $\omega_6:(\bar x_1\lor\bar x_2)$ was added to the formula,
it would not connect to any other clause in the RES graph because
the only clause containing~$x_1$ positively is $\omega_1=(x_1\lor x_2)$,
but that does not connect to~$\omega_6$ due to~$x_2$ appearing
negatively and positively in~$\omega_6$ and~$\omega_1$, respectively.
A similar type of analysis is done in
\emph{blocked clause elimination}~\cite{kullmann-dam99,jarvisalo-tacas10} --- a technique commonly used in formula preprocessing.

Although resolution-based graphs are not novel~\cite{yates-AI70} and
have been used in other domains~\cite{vangelder-cp11}, in this paper 
we propose to enhance the resolution-based graph representation by 
adding weights to edges.
Moreover, as far as we know, this representation has never been used 
for MaxSAT solving. 

\section{New Partition-based Algorithm for MaxSAT}
\label{sec:msu3-new}

Despite its very good performance in industrial partial MaxSAT instances,
the MSU3 algorithm (see Algorithm~\ref{alg:msu3}) may suffer from two
issues: (1) identification of unnecessarily large unsatisfiable subformulas
and, (2) a potentially large cardinality constraint to be maintained between
iterations. In fact these issues are related. If an unsatisfiable subformula
with an unnecessarily large number of soft clauses is encountered early, then 
an unnecessarily large cardinality constraint has to be dealt with through 
most of the algorithm's iterations.

Our approach to tackle these issues is to split the set of soft clauses.
The goal is that, at each iteration, the algorithm should only consider
part of the problem, instead of dealing with the whole problem instance
in each iteration.

\subsection{Algorithm Description}
\label{sec:msu3-new-desc}

\DontPrintSemicolon
\SetKwFunction{soft}{soft}
\SetKwFunction{SAT}{SAT}
\SetKwFunction{decomposeSoft}{partitionSoft}
\SetKwFunction{selectPart}{selectPartitions}
\SetKwFunction{first}{first}
\SetKwFunction{weight}{weight}
\SetKwFunction{encodeCNF}{CNF}
\SetKwFunction{min}{min}
\SetKwData{result}{satisfiable assignment to}
\SetKwData{unsat}{UNSAT}
\SetKwData{sat}{SAT}
\SetKwData{minc}{min$_\textnormal{c}$}
\SetKwData{true}{true}
\SetKwData{st}{st}
\SetVlineSkip{1pt}
\begin{algorithm}[!t]
  \small
  \KwIn{$\varphi = \varphi_h \cup \varphi_s$}
  \KwOut{satisfying assignment to $\varphi$}
  $\gamma \gets \langle \gamma_1, \ldots, \gamma_n\rangle \gets \decomposeSoft(\varphi_s, \varphi_h)$\label{li:dec:split}\;
  \ForEach{$\gamma_i \in \gamma$}{\label{li:msu3-begin}
    $(V_R^i, \lambda_i) \gets (\emptyset, 0)$\;
    $(\st, \varphi_C, \nu) \gets \SAT(\varphi_h \cup \gamma_i)$\;
    \While{$\st =$ \unsat}{
      \ForEach{$\omega \in (\varphi_C ~\cap~ \varphi_s)$}{
        $V_R^i \gets V_R^i \cup \{ r \}$\tcp*[r]{\footnotesize r is a new variable}
        $\omega_R \gets \omega \cup \{ r\}$\tcp*[r]{\footnotesize $\omega$ was not previously relaxed}
        $\gamma_i \gets (\gamma_i \setminus \{ \omega \}) \cup \{ \omega_R \}$\;
      }
      $\lambda_i \gets \lambda_i + 1$\;
      $(\st, \varphi_C, \nu) \gets \SAT(\varphi_h \cup \gamma_i \cup \{ \encodeCNF( \sum_{r\in V_R^i} r \leq \lambda_i ) \})$\label{li:msu3-end}\;
    }
  }
  \If{$|\gamma| = 1$}{
    \Return{$\nu$}\label{li:msu3Return}\tcp*[r]{\footnotesize no partitions were identified}
  }
  \While{\true}{
    $(\gamma_i, \gamma_j) \gets \selectPart(\gamma)$\label{li:selectPart}\;
    $\gamma \gets \gamma \setminus \{ \gamma_i, \gamma_j \}$\;
    $(\gamma_k, V_R^k, \lambda_k) \gets (\gamma_i \cup \gamma_j, V_R^i \cup V_R^j, \lambda_i+\lambda_j)$\label{li:initJoined}\;
    $(\st, \varphi_C, \nu) \gets \SAT(\varphi_h \cup \gamma_k \cup \{ \encodeCNF( \sum_{r\in V_R^k} r \leq \lambda_k ) \})$\label{li:msu3-K-begin}\;
    \While{$\st =$ \unsat}{
      \ForEach{$\omega \in (\varphi_C ~\cap~ \varphi_s)$}{
        $V_R^k \gets V_R^k \cup \{ r \}$\tcp*[r]{\footnotesize r is a new variable}
        $\omega_R \gets \omega \cup \{ r\}$\tcp*[r]{\footnotesize $\omega$ was not previously relaxed}
        $\gamma_k \gets (\gamma_k \setminus \{ \omega \}) \cup \{ \omega_R \}$\;
      }
      $\lambda_k \gets \lambda_k + 1$\;
      $(\st, \varphi_C, \nu) \gets \SAT(\varphi_h \cup \gamma_k \cup \{ \encodeCNF( \sum_{r\in V_R^k} r \leq \lambda_k ) \})$\label{li:msu3-K-end}\;
    }
    \If{$\gamma = \emptyset$}{
      \Return{$\nu$}\label{li:return}
    }
    \Else{
      $\gamma \gets \gamma \cup \{ \gamma_k \}$\label{li:add-K}\;
    }
  }
  \caption{Extended MSU3 Algorithm}\label{alg:msu3-part}
\end{algorithm}

Algorithm~\ref{alg:msu3-part} presents our enhancement of MSU3 with
partition of the soft clause set. The algorithm starts by partitioning
$\varphi_s$ into $n$ disjoint sets of soft clauses 
$\gamma_1, \gamma_2 \ldots \gamma_n$ (line~\ref{li:dec:split}). Observe
that several methods can be used to partition $\varphi_s$. Details
of this procedure are discussed later.

For each set $\gamma_i$, we apply the MSU3 algorithm to the formula 
$\varphi_h \cup \gamma_i$ (lines~\ref{li:msu3-begin}-\ref{li:msu3-end}). 
As a result, we obtain a lower bound value $\lambda_i$ associated with each 
set of soft clauses $\gamma_i$.
If the partitioning procedure creates a single partition, then the algorithm
terminates (line~\ref{li:msu3Return}). Otherwise, it is necessary to build
the solution of the MaxSAT instance by merging the different sets of soft
clauses.

The merge process works as follows. At each iteration, two sets of soft
clauses $\gamma_i$ and $\gamma_j$ are selected to be merged
(line~\ref{li:selectPart}) and removed from $\gamma$.
Let $\gamma_k$ denote the union of $\gamma_i$ and $\gamma_j$. 
Since $\gamma_i$ and $\gamma_j$ are disjoint, we necessarily have that
$\lambda_i + \lambda_j$ is a lower bound for $\gamma_k$. Hence, we
can safely initialize $\lambda_k = \lambda_i + \lambda_j$ (line~\ref{li:initJoined}).
Next, the lower bound $\lambda_k$ is refined by applying the MSU3 
algorithm to $\varphi_h \cup \gamma_k$ (lines~\ref{li:msu3-K-begin}-\ref{li:msu3-K-end}).
When set $\gamma$ becomes empty, then all soft clauses were merged
and the last solution found is an optimal solution (line~\ref{li:return}). 
Otherwise, there are still more sets to be merged and $\gamma_k$ is 
added to $\gamma$ (line~\ref{li:add-K}).

\subsection{Partition and Merge of Soft Clauses}
\label{sec:part-merge}

Algorithm~\ref{alg:msu3-part} can be configured differently depending
on two procedures: (1) how the set of soft clauses is partitioned 
(line~\ref{li:dec:split}) and (2) how to merge two sets of soft clauses 
(line~\ref{li:selectPart}).

In the partition procedure, our algorithm starts by representing the CNF
formula as a graph using one of the models described in section~\ref{sec:res}.
Next, we apply a community-finding algorithm on the graph representation
that maximizes a modularity measure~\cite{blondel08} in order to obtain
a graph partitioning.

Recently, the use of modularity measures has become widespread when
analyzing the structure of graphs, in particular for the identification
of communities~\cite{Girvan11062002,newman-girvan04}. In fact, this
has already been used in the analysis of SAT instances~\cite{ansotegui-sat12}
and to improve the initial unsatisfiability-based approach proposed
by Fu and Malik~\cite{FM06,martins-sat13}.
The purpose of the modularity measure is to evaluate the quality of
the partitions, where vertices inside a partition should be densely connected
and vertices assigned to different partitions should be loosely connected.
However, finding a set of partitions with an optimal modularity value is
computationally hard~\cite{brandes2006maximizing}. In our implementation,
we use the approximation algorithm proposed by Blondel et al.~\cite{blondel08}.

\begin{figure}[t]
  \centering
  \subfigure[Sequential merging]{
\begin{tikzpicture}[thick,xscale=.6,draw=black!70]
  \tikzstyle{every node}=[rectangle,draw=black!60,rounded corners=1mm,%
                          top color=white,bottom color=black!15,%
                          align=center,
                          minimum height=.5cm,minimum width=1cm,inner sep=2pt]
  \node(g1) at(0,0) {$\gamma_1$};
  \node(g2) at(2,0) {$\gamma_2$};
  \draw[dotted,very thick] (3.75,0)--(4.25,0);
  \node(gn) at(6,0) {$\gamma_{n}$};

  \node(gm1) at(1,1) {$\gamma_{1}\cup\gamma_{2}$};
  \draw(g1)--(gm1)--(g2);

  \node(gm2) at(2,2) {$\gamma_1\cup\dots\cup\gamma_{i}$};
  \draw[dotted,very thick](gm1)--(gm2)--(3,1);

  \node(gm3) at(3,3) {$\gamma_1\cup\dots\cup\gamma_{n}$};
  \draw[dotted,very thick](2.5,2.5)--(gm3.south);
  \draw(gm3)--(gn);
\end{tikzpicture}
    \label{fig:sequential}
  }
  \hspace{2em}
  \subfigure[Balanced merging]{
\begin{tikzpicture}[thick,xscale=.6,draw=black!70]
  \tikzstyle{every node}=[rectangle,draw=black!60,rounded corners=1mm,%
                          top color=white,bottom color=black!15,%
                          align=center,
                          minimum height=.5cm,minimum width=1cm,inner sep=2pt]
  \node(g1) at(0,0) {$\gamma_1$};
  \node(g2) at(2,0) {$\gamma_2$};
  \node(g3) at(1,1) {$\gamma_1\cup\gamma_2$};
  \draw(g1)--(g3)--(g2);
  \draw[dotted,very thick] (3.75,0)--(4.25,0);
  \node(gn1) at(6,0) {$\gamma_{n-1}$};
  \node(gn) at(8,0) {$\gamma_{n}$};
  \node(gm1) at(7,1) {$\gamma_{n-1}\cup\gamma_{n}$};
  \draw(gn)--(gm1)--(gn1);
  \node(gm3) at(4,3) {$\gamma_1\cup\dots\cup\gamma_{n}$};
  \draw[dotted,very thick] (2.5,2.5)--(gm3)--(5.5,2.5);
  \draw[dotted,very thick] (3.5,.8)--(4,1.3)--(4.5,.8);
  \draw[dotted,very thick] (3.2,1.8)--(4,2.3)--(4.8,1.8);
\end{tikzpicture}
    \label{fig:weighted}
  }
  \caption{Examples of merge processes}
\end{figure}
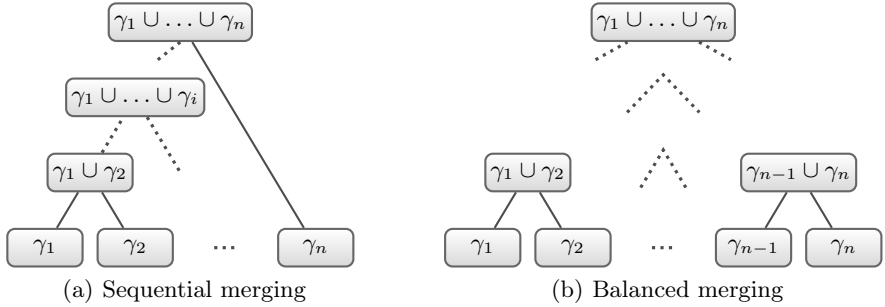

At each iteration in Algorithm~\ref{alg:msu3-part}, two partitions
are selected to be merged. One can devise several different criteria
to select and merge the partitions of soft clauses.
In early attempts, the merge process was sequential~\cite{martins-sat13}.
Given $n$ partitions $\gamma_1, \gamma_2 \ldots \gamma_n$, at iteration $i$ ($i \le n$) 
of the algorithm, the first $i$ partitions 
$\gamma_1, \gamma_2 \ldots \gamma_i$ were merged sequentially.

Figure~\ref{fig:sequential} illustrates the sequential
merging procedure.
Observe that the sequential merging process is not balanced.
This results in an early growth of the identified subformulas
and, as a result, an early growth of the cardinality constraints
to be maintained at each iteration of the algorithm.

In this paper, we propose a weighted balanced merge procedure
that depends on the strength of the graph connections between partitions.
The goal is to delay having to deal with a large number of
soft clauses, until the latter iterations of the algorithm.
Figure~\ref{fig:weighted} illustrates the weighted balanced
merging procedure.

Let $G = (V, E)$ denote an undirected weighted graph where $V$
is the set of vertices and $E$ the set of edges.
Let $w : E \rightarrow \mathbb{R}$ be a weight function for each edge in 
the graph.
The community-finding algorithm identifies a set of communities
$C = \{ C_1, C_2, \ldots, C_n \}$ where every vertex $u \in V$
is assigned to one and only one community in $C$. 
Hence, since in both CVIG and RES model there is a node for each
propositional clause, one can build the partitions in a straightforward
manner. For each community $C_i$ with vertices representing soft clauses,
there is a partition $\gamma_i$ containing the respective soft clauses.

Based on the graph representation, one can define the strength of the
connection between partitions. Let $d_{ij}$ denote the strength between
partition $\gamma_i$ and $\gamma_j$. One can define $d_{ij}$ based on the
weight between the vertices of their respective communities $C_i$ and 
$C_j$ in the graph. Hence, $d_{ij}$ can be defined as follows:
\begin{equation}
  d_{ij} = \sum_{u \in C_i \wedge v \in C_j} w(u, v)
\end{equation}
Considering that the graph is undirected, we necessarily have
that $d_{ij} = d_{ji}$.

Given an initial set $\gamma$ of $n$ partitions 
$\gamma_1, \gamma_2 \ldots, \gamma_n$, our algorithm applies a 
greedy procedure that pairs all partitions $\gamma_i$ and $\gamma_j$ 
from $\gamma$ to be merged, starting with the pair with largest 
$d_{ij}$. After pairing all partitions in the initial set, we 
perform the same procedure to the next $n/2$ partitions that 
result from the initial merging iterations. 
This is iteratively applied until we only have a single partition
(see Figure~\ref{fig:weighted}).

Observe that if partitions $\gamma_i$ and $\gamma_j$ are merged 
into a new partition $\gamma_k$, then the connectivity strength 
$d_{kl}$ between $\gamma_k$ to another partition $\gamma_l$ is 
given by $d_{kl} = d_{il}+d_{jl}$. 
This follows from the fact that the communities in the graph 
are disjoint.

Finally, we would like to reference other solvers that split the set 
of soft clauses by identifying disjoint unsatisfiable 
subformulas~\cite{msuncore-aaai11,morgado-sat12}. However, there are major 
differences with regard to our proposed approach. First, our solver takes 
advantage of an explicit formula representation to split the set of soft 
clauses, instead of using the unsatisfiable subformulas provided by the SAT 
solver. Moreover, in our solver, the merge process is also guided by the 
explicit representation of the formula.

Furthermore, in solvers where disjoint unsatisfiable subformulas are
identified~\cite{msuncore-aaai11,morgado-sat12}, the split
occurs on the cardinality constraints at each iteration. However,
each SAT call still has to deal with the whole formula at
each iteration.
In Algorithm~\ref{alg:msu3-part}, the SAT solver does not have 
to deal with all soft clauses at each iteration, but only after
the final merge step.

\subsection{Algorithm Analysis}
\label{sec:analysis}

In this section a proof sketch of the correctness Algorithm~\ref{alg:msu3-part}, 
as well as an analysis on the number of SAT calls is presented.

\begin{proof}
As mentioned in section~\ref{sec:prelim}, we assume the set of hard clauses $\varphi_h$
is satisfiable. Otherwise, the MaxSAT formula is unsatisfiable. This can be verified
by a single SAT call on $\varphi_h$ before applying Algorithm~\ref{alg:msu3-part}.

For the proof we adopt the following notation. For some set $\gamma_i$ processed in
Algorithm~\ref{alg:msu3-part}, we write $\gamma_i^R\subseteq\varphi_s$ for the set of clauses that
were relaxed in the algorithm (but clauses in $\gamma_i^R$ do not contain the relaxation variables).  We will
prove by induction the invariant that $\varphi_h\cup\gamma_i^R$ cannot be satisfied unless at
least $\lambda_i$ clauses are removed from $\gamma_i^R$.  The induction hypothesis is satisfied
trivially at the beginning of the algorithm as each $\lambda_i$ is initialized to~$0$.

Consider the case where $\lambda_i$ is augmented by $1$ when
$\varphi_h\cup\gamma_i\cup\{\sum_{r\in V_R^i}r\leq\lambda_i\}$ is unsatisfiable.
Let~$\varphi_C$ be the obtained unsatisfiable subformula from the SAT call, 
let $\varphi_C^R\subseteq\varphi_s$ be the soft clauses of $\varphi_C$ that appear 
as relaxed in $\gamma_i$ and let $\varphi_C^N=\varphi_s\cap\varphi_C$ be the rest of 
the soft clauses in the unsatisfiable subformula (not yet relaxed).  
From induction hypothesis 
$\varphi_h\cup\varphi_C^R$ cannot be satisfied unless at least
$\lambda_i$ clauses are removed from $\varphi_C^R\subseteq\gamma_i^R$.
Since $\varphi_C$ is an unsatisfiable subformula, it is \emph{impossible} to satisfy
$\varphi_h\cup\varphi_C^R\cup\varphi_C^N$
by removing $\lambda_i$ clauses from~$\varphi_C^R$.
Now we need to also show that it is impossible to satisfy
$\varphi_h\cup\gamma_i^R\cup\varphi_C^N$ by removing
$\lambda_i$ clauses from $\gamma_i^R\cup\varphi_C^N$ (this is the new set of relaxed clauses).

Let us assume for contradiction that it is possible to satisfy
$\gamma_i^R\cup\varphi_C^N$ by removing some set of clauses $\xi$ s.t.\ $|\xi|=\lambda_i$.
To show the contradiction we consider two cases:
(1)~$\xi\subseteq\gamma_i$ and
(2)~$\xi\nsubseteq\gamma_i$.
Case~(1) yields an immediate contradiction
as we would have not obtained unsatisfiability in the SAT call as it would be possible
to satisfy $\varphi_h\cup\gamma_i^R$ by removing $\lambda_i$ clauses from~$\gamma_i^R$.
For case~(2) consider that there is a clause $\omega\in\xi$ s.t.\ $\omega$ is not yet relaxed,
i.e.\ $\omega\notin\gamma^R$.
This means that $\varphi_h\cup\gamma_i^R$
is satisfiable after removing \emph{less} than
$\lambda_i$ clauses, which is a contradiction with the induction hypothesis.

To show that the invariant is preserved by the merge operation,
we observe that any merged $\gamma_i$ and $\gamma_j$ are disjoint
and therefore so are $\gamma_i^R$ and $\gamma_j^R$.
In order to satisfy $\varphi_h\cup(\gamma_i^R\cup\gamma_j^R)$,
both $\varphi_h\cup\gamma_i^R$,
     $\varphi_h\cup\gamma_j^R$ must be satisfied.
Consequently, at least $\lambda_i+\lambda_j$ clauses must be removed
from $(\gamma_i^R\cup\gamma_j^R)$.

\qed
\end{proof}

Finally, we note that the number of SAT calls performed by Algorithm~\ref{alg:msu3-part}
is larger than the MSU3 algorithm. Observe that the number of unsatisfiable
SAT calls is the same for both algorithms. Let $\lambda$ be the number of
unsatisfiable soft clauses at any optimal solution of the MaxSAT instance.
In this case, both algorithms perform $\lambda$ unsatisfiable SAT calls.
However, while MSU3 performs only one satisfiable SAT call, 
Algorithm~\ref{alg:msu3-part} performs $2n-1$, where $n$ is the number of 
identified partitions (line~\ref{li:dec:split}).

\section{Experimental Results}
\label{sec:results}

In this section we compare different configurations of 
Algorithm~\ref{alg:msu3-part} with the top 3 non-portfolio solvers 
of the MaxSAT 2014 Evaluation's partial MaxSAT category.
The top 3 were Open-WBO's MSU3 incremental algorithm~\cite{martins-sat14,martins-cp14}, 
Eva500a~\cite{nina-aaai14} and MSCG~\cite{morgado-cp14}. 
The new partition-based algorithm is also implemented using
the Open-WBO framework\footnote{Available at \url{http://sat.inesc-id.pt/open-wbo/}}.

The algorithms were evaluated running on the power set of the partial MaxSAT 
industrial instances of the MaxSAT evaluations of 2012, 2013 and 2014. 
For each instance, algorithms were executed with a timeout of 1800 seconds 
and a memory limit of 4 GB. 
Similar resource limitations were used during the last MaxSAT Evaluation of 2014. 
These tests were conducted on a machine with 
4 AMD Opteron 6376 (2.3 GHz) and 128 GB of RAM, running Debian jessie.

\begin{table*}[!t]
\centering
\scriptsize
\begin{tabular} { l || r || r | r | r | r | r | r | r }
\textbf{Instance Group} & \textbf{Total} & \textbf{MSU3} & \textbf{Eva500a} & \textbf{MSCG} & \textbf{S-CVIG} & \textbf{S-RES} & \textbf{W-CVIG} & \textbf{W-RES} \\
  \hline
aes & 7 & 1 & 1 & 1 & 1 & 1 & 1 & 1 \\
atcoss/mesat & 18 & 11 & 11 & 4 & 11 & 1 & 11 & 11 \\
atcoss/sugar & 19 & 12 & 11 & 4 & 12 & 3 & 12 & 12 \\
bcp/fir & 59 & 59 & 55 & 59 & 56 & 44 & 51 & 51 \\
bcp/hipp-yRa1/simp & 17 & 16 & 16 & 16 & 16 & 16 & 16 & 16 \\
bcp/hipp-yRa1/su & 38 & 35 & 34 & 33 & 34 & 34 & 35 & 33 \\
bcp/msp & 64 & 26 & 37 & 29 & 23 & 41 & 27 & 42 \\
bcp/mtg & 40 & 40 & 40 & 40 & 40 & 40 & 40 & 40 \\
bcp/syn & 74 & 43 & 48 & 47 & 47 & 48 & 46 & 49 \\
circuit-trace-compaction & 4 & 4 & 4 & 4 & 4 & 3 & 4 & 4 \\
close-solutions & 50 & 48 & 48 & 46 & 40 & 32 & 40 & 45 \\
des & 50 & 42 & 41 & 41 & 49 & 48 & 50 & 48 \\
haplotype-assembly & 6 & 5 & 5 & 5 & 5 & 5 & 5 & 5 \\
hs-timetabling & 2 & 1 & 1 & 0 & 1 & 1 & 1 & 1 \\
mbd & 46 & 45 & 42 & 43 & 44 & 45 & 45 & 45 \\
packup-pms & 40 & 40 & 40 & 40 & 40 & 40 & 40 & 40 \\
pbo/mqc/nencdr & 84 & 84 & 84 & 84 & 84 & 84 & 84 & 84 \\
pbo/mqc/nlogencdr & 84 & 84 & 84 & 84 & 84 & 84 & 84 & 84 \\
pbo/routing & 15 & 15 & 15 & 15 & 14 & 15 & 15 & 15 \\
protein\_ins & 12 & 12 & 8 & 12 & 12 & 12 & 12 & 12 \\
tpr/Multiple\_path & 48 & 48 & 44 & 42 & 48 & 48 & 48 & 48 \\
tpr/One\_path & 50 & 50 & 50 & 50 & 50 & 50 & 50 & 50 \\
\hline
Total & 827 & 721 & 719 & 699 & 715 & 695 & 717 & 736 \\
\end{tabular}
\vspace{1em}
\caption[Experimental evaluation of Open-WBO's MSU3 algorithm, Eva500a, MSCG and 4 different configurations of the partition-based algorithm]
{Experimental evaluation of Open-WBO's MSU3 algorithm, Eva500a, MSCG and 4 different configurations of the partition-based algorithm.}
\label{table:results}
\end{table*}

Table \ref{table:results} presents the number of instances solved by each 
algorithm, per instance set. Besides MSU3, Eva500a and MSCG, results for
the best 4 configurations of the partition-based enhanced MSU3 algorithm 
are shown.
S-CVIG applies the sequential merging of partitions using the CVIG graph model.
S-RES also applies sequential merging, but using the RES graph model.
W-CVIG and W-RES apply the weighted balanced merging of partitions, using 
the CVIG and RES graph models, respectively. Note that all our implementations
are fully incremental, i.e. only one instance of the SAT solver is
created throughout the execution of the proposed algorithm. As with 
the MSU3 implementation on Open-WBO, we take advantage of assumptions 
usage at each SAT call and incremental encoding of cardinality 
constraints~\cite{martins-cp14}.

Results from Table \ref{table:results} show that all variants of the 
partition-based algorithm are competitive with the remaining state of 
the art algorithms. However, overall results clearly show that W-RES 
outperforms all remaining algorithms, since it is able to solve
more instances in total.
Moreover, results for the configurations of partition-based algorithm 
also show that weight-based balanced merging of partitions is preferable 
to sequential partitioning.

Considering that MSU3 is our base solver, most gains occur in instance 
sets {\tt bcp/msp}, {\tt bcp/syn} and {\tt des}. While in the 
{\tt bcp/syn} and {\tt des} instance sets, all partition-based configurations
perform better, in {\tt bcp/msp} the resolution-based graph partitioning
allowed a significant performance boost. 

\begin{figure}[t]
  \centering
  \subfigure[{\tt des} instance set]{
    \includegraphics[width=0.45\textwidth]{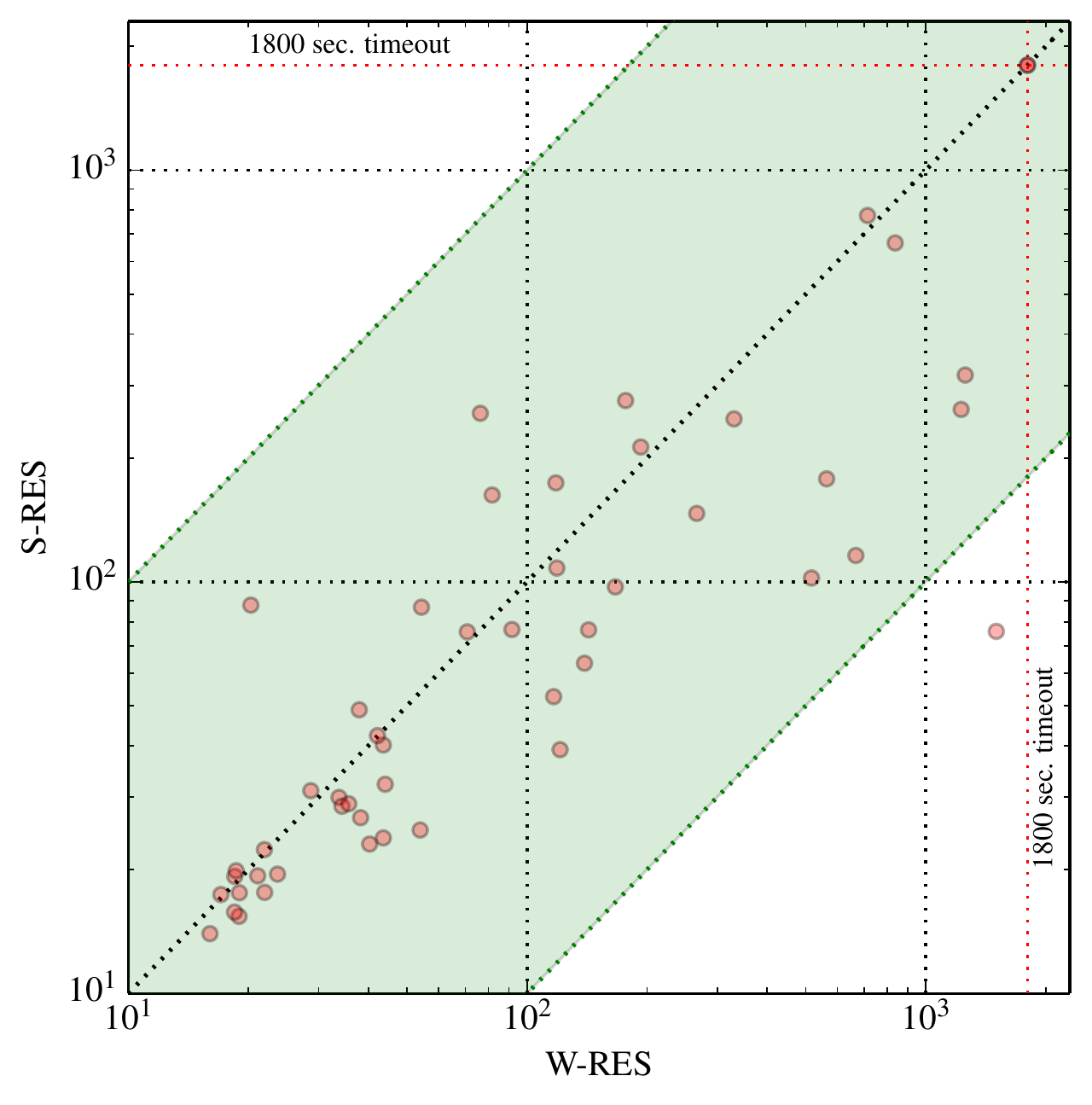}
    \label{fig:scatter-sortres-binres-des}
  }
  \quad
  \subfigure[{\tt bcp/msp} instance set]{
    \includegraphics[width=0.45\textwidth]{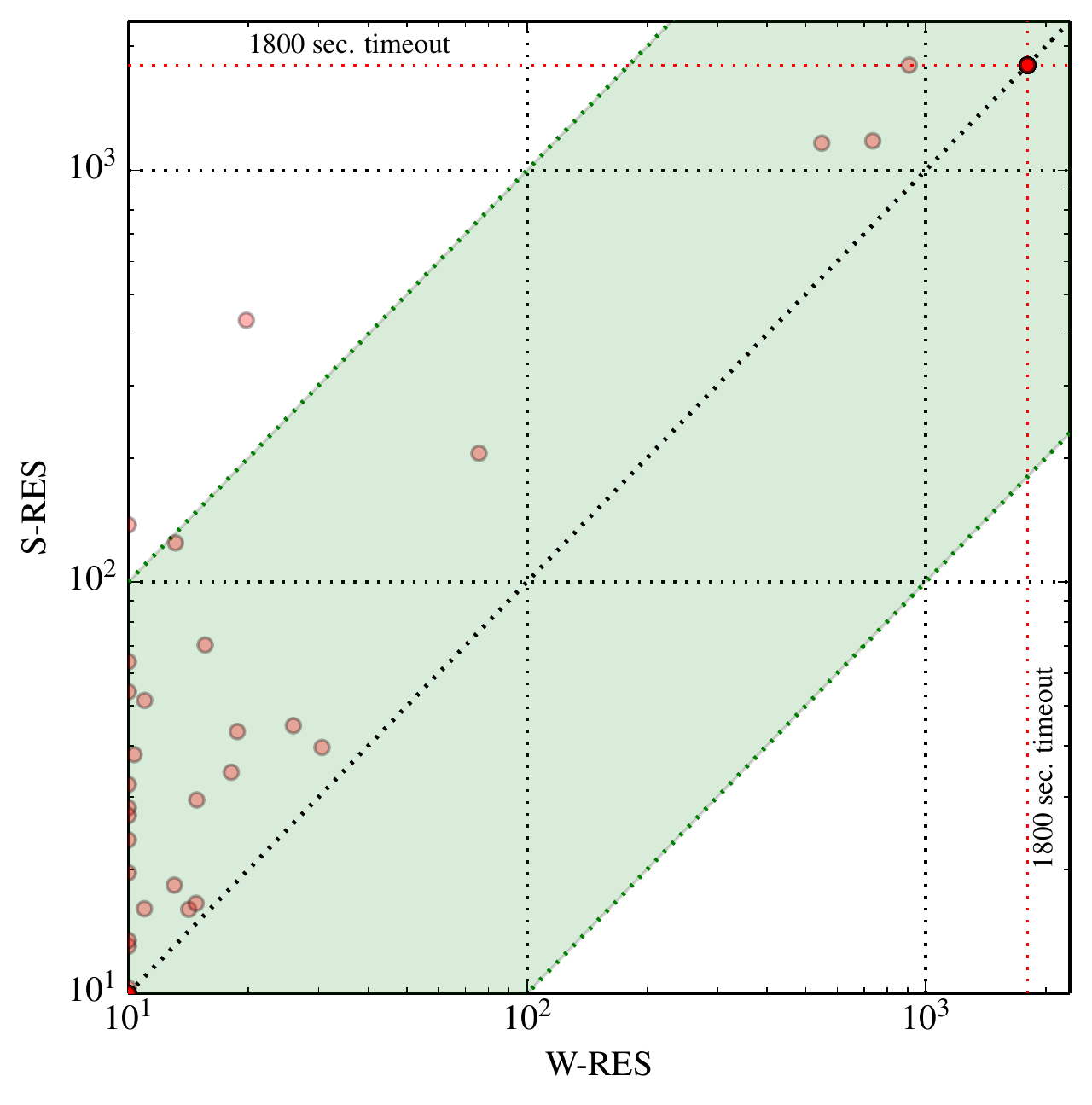}
    \label{fig:scatter-sortres-binres-msp}
  }
  \caption{Comparison between run times of S-RES and W-RES on {\tt des} and {\tt bcp/msp} instance sets}
\end{figure}

Figures~\ref{fig:scatter-sortres-binres-des} and~\ref{fig:scatter-sortres-binres-msp} 
compare the results of S-RES and W-RES on the {\tt des} and {\tt bcp/msp} instance sets. 
In the {\tt des} instances, the run time of sequential merging is slightly better,
despite solving the same number of instances. Nevertheless, in the {\tt bcp/msp} 
instance set the weight-based balanced merging used in W-RES clearly outperforms 
the sequential merging approach used in S-RES. 

\begin{figure}[t]
  \centering
  \subfigure[{\tt des} instance set]{
    \includegraphics[width=0.45\textwidth]{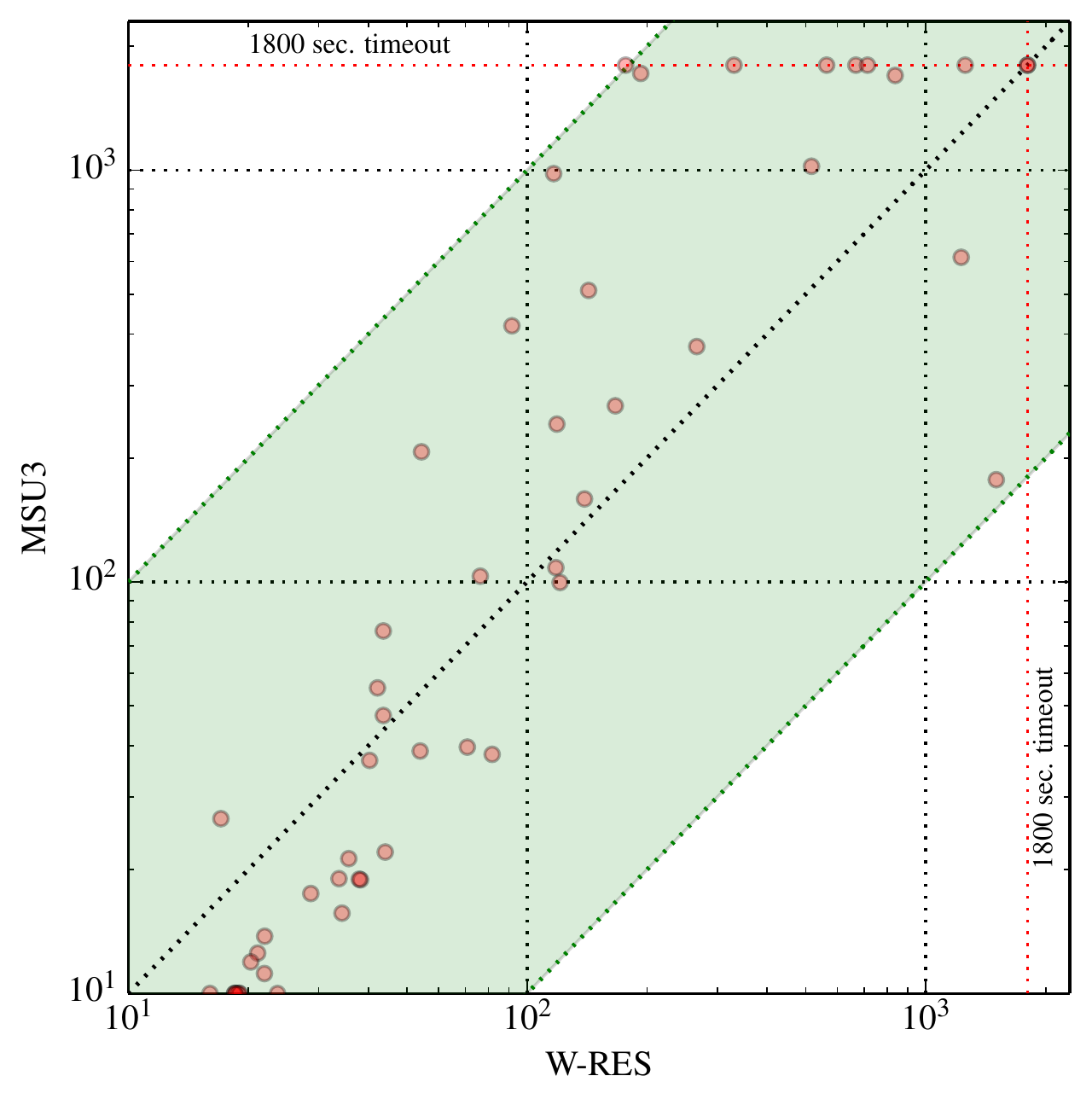}
    \label{fig:scatter-msu3-binres-des}
  }
  \quad
  \subfigure[{\tt bcp/msp} instance set]{
    \includegraphics[width=0.45\textwidth]{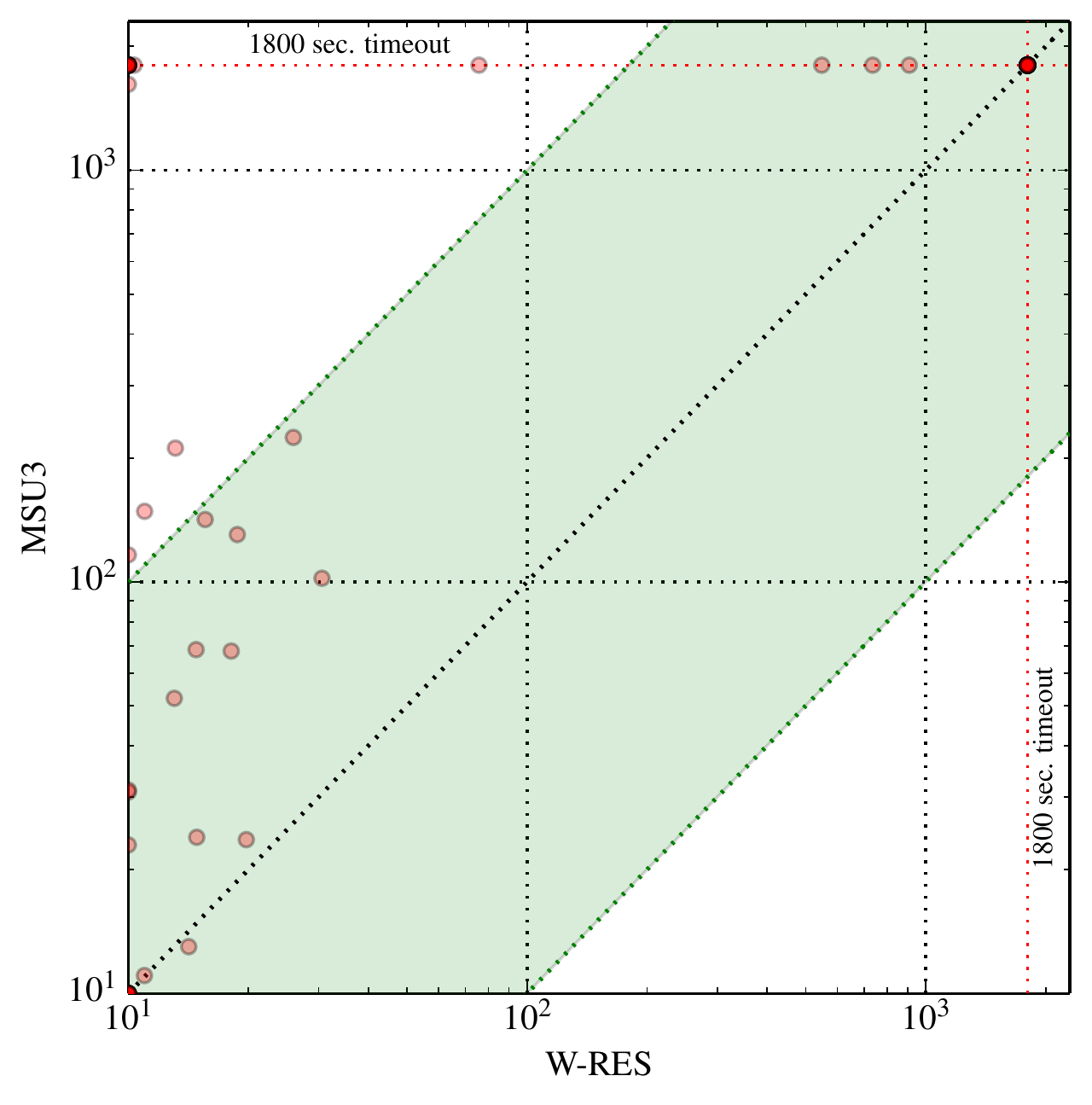}
    \label{fig:scatter-msu3-binres-msp}
  }
  \caption{Comparison between run times of MSU3 and W-RES on {\tt des} and {\tt bcp/msp} instance sets}
\end{figure}

In Figures~\ref{fig:scatter-msu3-binres-des} and~\ref{fig:scatter-msu3-binres-msp}
we compare MSU3 and W-RES on the same benchmark sets. It can be observed that
W-RES performs much better in these instances. In the {\tt des} instance set,
there are some instances where W-RES is not as fast, since there is some
time spent in finding partitions and additional SAT calls. 
We note that there is always some time spent in building the graph,
applying the community finding algorithm and splitting the set of soft clauses. 
However, this partitioning step is usually not very time consuming.
Nevertheless, W-RES is able to scale better and solve more instances.
In the {\tt bcp/msp} instances, the proposed techniques allow W-RES 
to be much better than MSU3, as well as all other algorithms tested.

Resolution-based graph models performed worst in the {\tt bcp/fir} 
category. It was observed that the overall modularity values obtained 
for the resolution-based graphs were low in this particular instance set. 
As a result, the partitioning obtained for S-RES and W-RES
in {\tt bcp/fir} instances is not as meaningful as for other instance sets.
Hence, when this occurs, it can deteriorate the algorithm's performance,
since the partition-based algorithm performs more SAT calls than MSU3.

\begin{figure}[!t]
\centering
\includegraphics[width=1\textwidth]{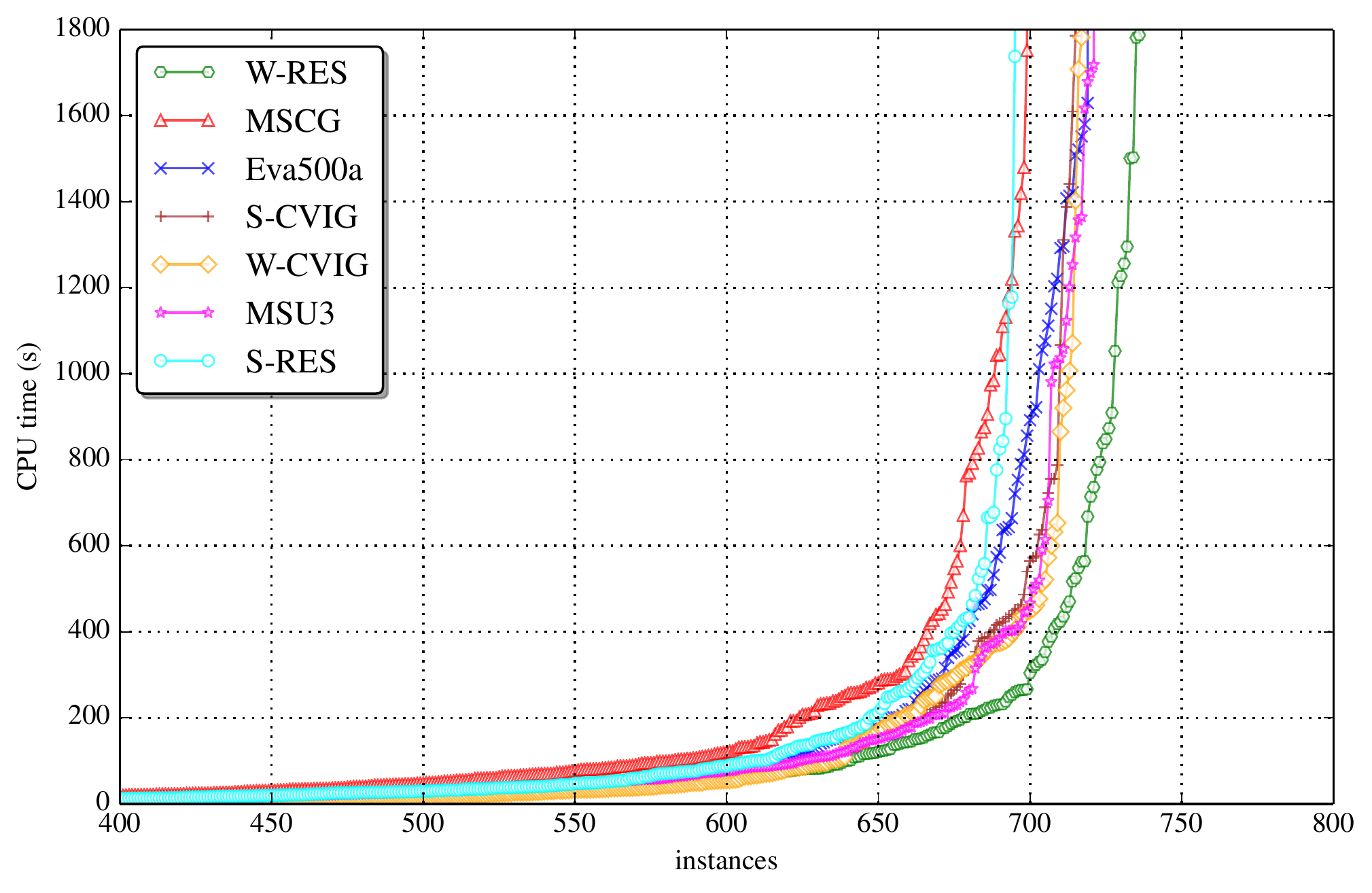}
\caption[Run times cactus]{Cactus plot with the run times for MSU3, Eva500a, MSCG, S-CVIG, S-RES, W-CVIG and W-RES.}
\label{fig:cactus}
\end{figure}

Finally, Figure \ref{fig:cactus} shows a cactus plot with the run times of all 
algorithms considered in the experimental evaluation. Here we can observe that
S-RES is much slower than W-RES, clearly showing the effectiveness of the
newly proposed weight-based merging. Overall, W-RES clearly outperforms 
the remaining algorithms, being able to solve 700 instances in 300 seconds or less.



\section{Conclusions and Future Work}
\label{sec:conc}

In this paper we exploit resolution-based graph representations of CNF
formulas in order to develop a new state of the art algorithm 
for MaxSAT. In the proposed approach, soft clauses are initially 
partitioned in disjoint sets by analyzing the formula structure.
The partitioning process is attained by applying a community-finding 
algorithm on weighted resolution-based graphs. Next, at each iteration
of the algorithm, partitions are merged using structural information from 
the graph representation until an optimal solution is found.

The proposed approach is novel in many aspects. First, the use of
a resolution-based graph representation allows to better model the
interaction between clauses. Furthermore, instead of applying a
sequential merging process, the graph representation is also
used in a weight-based balanced merging procedure. Moreover, since 
the algorithm does not have to deal with the whole formula at each
iteration, smaller unsatisfiable cores are identified. As a
result from this process, smaller cardinality constraints
are encoded into CNF at each iteration, thus improving the
algorithm's performance.

Experimental results obtained in industrial partial MaxSAT
instances clearly show the effectiveness of the proposed
algorithm. As a result, our solver improves upon the best
non-portfolio solver from the 2014 MaxSAT solver evaluation.

The source code of the new solver will become available as part 
of the Open-WBO framework. This will allow the research community
to build upon the current work to further improve MaxSAT solving.

As future work, we propose to extend the proposed approach for
weighted MaxSAT solving. Moreover, different model representations
of CNF formulas are to be tested, as well as new techniques for
building and merging partitions of soft clauses in MaxSAT formulas.
Furthermore, the proposed techniques are not exclusive to MSU3
and can also be integrated into other MaxSAT algorithms. Additionally,
these techniques can also be applied to other extensions of SAT.

\section*{Acknowledgments}
This work was partially supported by the ERC project 280053, 
FCT grant POLARIS (PTDC/EIA-CCO/123051/2010), 
FCT grant AMOS (CMUP-EPB/\-TIC/\-0049/\-2013),
and INESC-ID's multiannual PIDDAC funding PEst-OE/\-EEI/\-LA0021/2013.

\bibliography{sat15}
\bibliographystyle{splncs03}

\end{document}